\documentclass[sigconf]{acmart}
\AtBeginDocument{%
  }

\usepackage{dsfont}   
\usepackage{microtype}
\usepackage{multirow}

\setcopyright{acmlicensed}
\copyrightyear{2026}
\acmYear{2026}
\acmDOI{XXX.XXXXXXX.XXXXXXX}
\acmConference[ACM MM '26]{Proceedings of the 34th ACM International Conference on Multimedia}{November 10--14, 2026}{Rio de Janeiro, Brazil}
\acmBooktitle{Proceedings of the 34th ACM International Conference on Multimedia (ACM MM '26)}
\acmISBN{978-1-4503-XXXX-X/2026/04}

\begin{document}

\title{Agri-R1: Agricultural Reasoning for Disease Diagnosis via Automated-Synthesis and Reinforcement Learning}

\author{Wentao Zhang}
\affiliation{%
  \institution{Shandong University of Technology}
  \city{Shandong}
  \country{China}
}
\email{24418011424@stumail.sdut.edu.cn}

\author{Mingkun Xu}
\affiliation{%
  \institution{Guangdong Institute of Intelligence Science and Technology}
  \city{Hengqin, Zhuhai}
  \country{China}}
\email{xunmingkun@gdiist.cn}

\author{Qi Zhang}
\affiliation{%
  \institution{Faculty of Data Science, City University of Macau}
  \city{Macau SAR}
  \country{China}
}
\email{qizhang@cityu.edu.mo}

\author{Shangyang Li}
\affiliation{%
 \institution{School of Physical Science and Technology, Beijing University of Posts and Telecommunications}
 \city{Beijing}
 \country{China}}
\email{nic_lab@163.com}

\author{Derek F. Wong}
\affiliation{%
  \institution{NLP2CT Lab, Department of Computer and Information Science, University of Macau}
  \city{Macau SAR}
  \country{China}}
  \email{derekfw@um.edu.mo}

\author{Lifei Wang}
\affiliation{%
  \institution{Institute of International Language Services Studies, Macau Millennium College}
  \city{Macau SAR}
  \country{China}}
\email{wanglifei@mmc.edu.mo}

\author{Yangchao Yang}
\affiliation{%
  \institution{Institute of International Language Services Studies, Macau Millennium College}
  \city{Macau SAR}
  \country{China}}
\email{yangyanchao@mmc.edu.mo}

\author{Lina Lu}
\authornotemark[1]
\affiliation{%
  \institution{Shandong University of Technology}
  \city{Shandong}
  \country{China}}
\email{think0759@sdut.edu.cn}

\author{Tao Fang}
\authornote{Co-corresponding Author}
\authornote{Tao Fang is the Senior Corresponding Author (Last Author) 
            and the principal supervisor of this work. 
            He led the research design, guided the methodology, 
            and oversaw the entire project.}
\affiliation{%
  \institution{Institute of International Language Services Studies, Macau Millennium College}
  \city{Macau SAR}
  \country{China}}
\email{taofang@mmc.edu.mo}
\renewcommand{\shortauthors}{Trovato et al.}

\begin{abstract}
Agricultural disease diagnosis challenges VLMs, as conventional fine-tuning requires extensive labels, lacks interpretability, and generalizes poorly. While reasoning improves model robustness, existing methods rely on costly expert annotations and rarely address the open-ended, diverse nature of agricultural queries.
To address these limitations, we propose \textbf{Agri-R1}, a reasoning-enhanced large model for agriculture.
Our framework automates high-quality reasoning data generation via vision-language synthesis and LLM-based filtering, using only 19\% of available samples.
Training employs Group Relative Policy Optimization (GRPO) with a novel reward function that integrates domain-specific lexicons and fuzzy matching to assess both correctness and linguistic flexibility in open-ended responses.
Evaluated on CDDMBench, our resulting 3B-parameter model achieves performance competitive with 7B- to 13B-parameter baselines, showing a +27.9\% relative gain in disease recognition accuracy, +33.3\% in agricultural knowledge QA, and a +26.10-point improvement in cross-domain generalization over standard fine-tuning. These results suggest that automated reasoning synthesis paired with domain-aware reward design may provide a broadly applicable paradigm for RL-based VLM adaptation in data-scarce specialized domains. Our code and data are publicly available at: \textcolor{blue}{\url{https://github.com/CPJ-Agricultural/Agri-R1}}.
\end{abstract}




\maketitle

\section{Introduction}
\label{sec:introduction}

Agricultural crop diseases pose a persistent threat to global food security, causing substantial yield losses and economic damage~\cite{savary2020modeling,gai2024plant,shahbazi2025losses}. Accurate and timely diagnosis is essential for effective crop protection, yet remains challenging due to complex visual symptoms and limited expert availability in many regions~\cite{upadhyay2025deep,ngugi2024revolutionizing,buja2021advances,mohanty2016using}. Recent advances in Vision-Language Models (VLMs) have demonstrated significant promise for automated diagnosis through visual question answering (VQA), enabling farmers to submit crop images accompanied by natural language queries to obtain diagnostic guidance~\cite{lu2024application,sapkota2025multi}.



The predominant paradigm for adapting VLMs to agricultural tasks is SFT. Although effective within the training domain, SFT suffers from three critical limitations that hinder its real-world deployment. First, it is data-hungry, requiring massive labeled datasets that are prohibitively expensive to obtain in resource-constrained agricultural settings~\cite{liu2024multimodal}. Second, it offers limited interpretability, as models produce diagnostic labels without explicating their underlying reasoning. This "black-box" behavior undermines farmer trust and prevents effective validation by agricultural extension agents~\cite{zhi2025medgr,chu2025sft}. Third, it generalizes poorly, as models tend to memorize dataset-specific patterns rather than acquire robust diagnostic reasoning. Consequently, their performance deteriorates sharply under domain shifts, such as encountering new crops, varying lighting conditions, or concurrent infections~\cite{pan2025medvlm,wu2023laboratory,nanavaty2024integrating,chen2025sft}. Collectively, these limitations reveal a fundamental gap: the need for models that are not only accurate but also data-efficient, interpretable, and robust to the open-ended diversity inherent in real-world agricultural queries.

Structured reasoning enhances model transparency by generating explicit intermediate reasoning steps, while reinforcement learning (RL) presents a promising alternative to SFT by promoting diverse reasoning strategies through reward guidance~\cite{shakya2023reinforcement}. GRPO~\cite{shao2024deepseekmath,wang2025grpo,tong2025delving} has achieved strong generalization in mathematical and coding tasks via group-based advantage estimation. However, a direct application in agriculture faces two synergistic bottlenecks. First, constructing high-quality CoT data is prohibitively expensive, requiring domain experts to manually annotate reasoning chains. Second, existing RL applications in medical VQA~\cite{yi2022automated,hu2023reinforcement} primarily target \emph{closed-set multiple-choice questions} with binary rewards.
This paradigm is fundamentally mismatched with agricultural VQA, which requires evaluating open-ended, linguistically diverse responses for both factual correctness and reasoning quality---a challenge that remains unaddressed in prior work.

To overcome these bottlenecks, we introduce \textbf{Agri-R1}, to our knowledge the first GRPO-based framework designed specifically for open-ended, reasoning-enhanced agricultural VQA. We integrate three key innovations to simultaneously achieve data efficiency, interpretability, and robustness: (1) we eliminate manual CoT annotation costs through an automated pipeline that synthesizes reasoning chains via VLMs and filters high-quality data using LLM-as-a-Judge, constructing a compact yet powerful dataset from only 19\% of the original corpus; (2) to address the unique challenge of evaluating open-ended answers, we construct agricultural domain vocabularies and design a novel fuzzy-matching reward function. This function assesses not just correctness but also the linguistic appropriateness of responses, enabling effective policy optimization far beyond binary rewards; (3) we demonstrate that GRPO-driven policy optimization, fueled by our automated reasoning data and specialized reward, enables a remarkably compact \textbf{3B-parameter model} to achieve superior accuracy and cross-domain generalization compared to significantly larger baselines trained on full datasets.

Our primary contributions, findings, and results are as follows:
\begin{itemize}
\item We propose \textbf{Agri-R1}, to our knowledge the first GRPO-based framework specifically designed for agricultural disease diagnosis. It introduces a fully automated pipeline that synthesizes and filters high-quality Chain-of-Thought reasoning data without requiring any expert annotations, using only 19\% of the original samples.
\item We design a novel domain-aware reward mechanism based on agricultural lexicons and multi-tier fuzzy matching. This mechanism jointly evaluates both factual correctness and linguistic flexibility in open-ended responses, effectively overcoming the critical limitations of binary-reward systems.
\item We demonstrate that a compact 3B-parameter model trained with our framework significantly outperforms much larger (7B--13B) supervised fine-tuning baselines across disease recognition accuracy (+27.9\% relative gain), agricultural knowledge QA (+33.3\% relative gain), and cross-domain generalization (+26.10 points on AgMMU), highlighting the powerful synergy between automated reasoning synthesis and reinforcement learning exploration.
\item Through detailed analysis (Sections \ref{sec:bias_analysis} and \ref{sec:disease_analysis}), we identify frequency-induced gradient competition as a key failure mode that degrades performance on rare crops and diseases under standard GRPO, revealing important insights for improving long-tail robustness in future RL-based VLM adaptation.
\end{itemize}


\begin{figure*}[t]
\centering
\includegraphics[width=\textwidth]{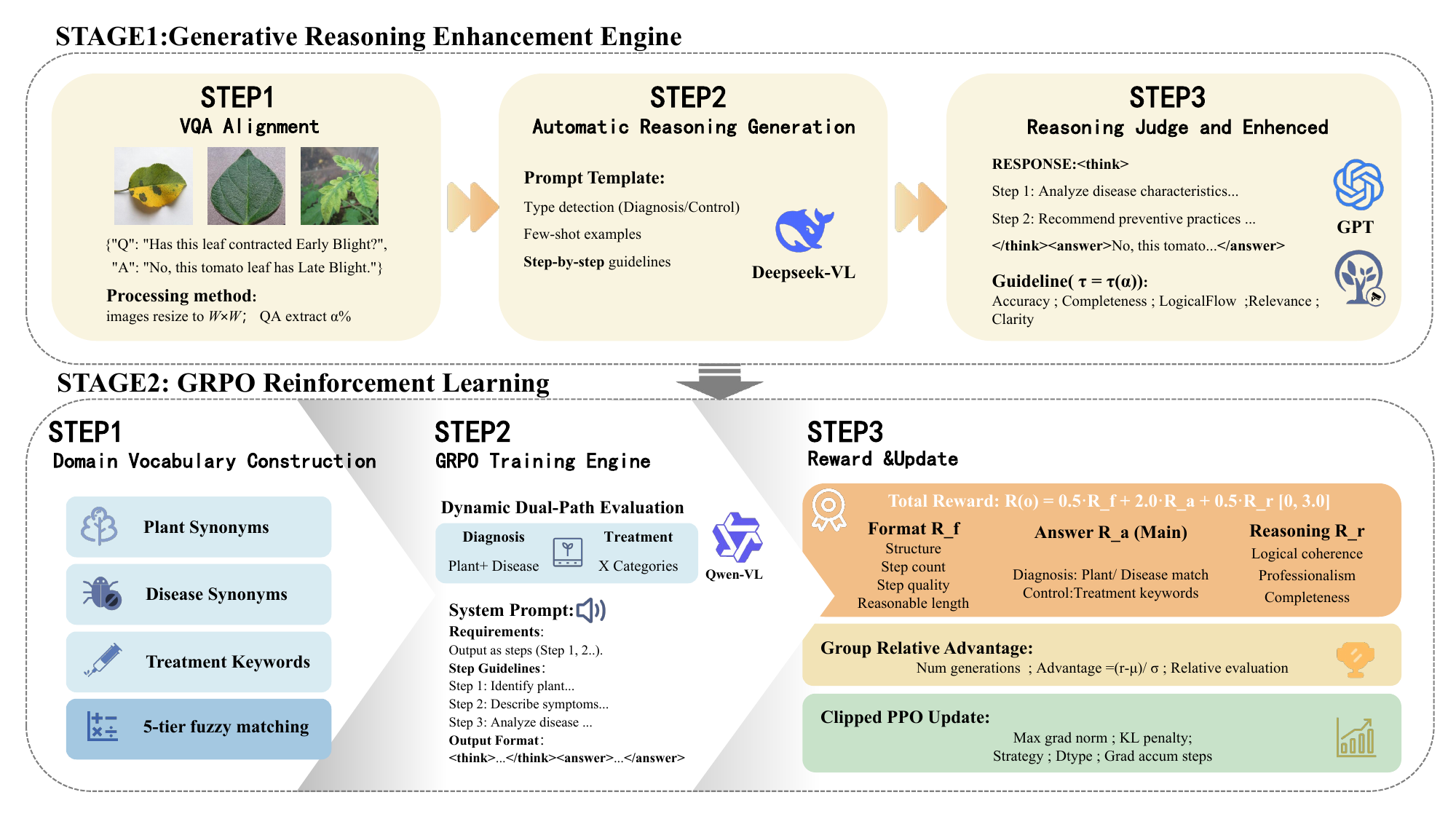}
\caption{
Overview of our proposed two-stage GRPO framework. In Stage 1, raw VQA pairs are transformed into reasoning exemplars: a SOTA VLM generates structured reasoning chains, which are subsequently filtered by an LLM-based judge to ensure quality. Stage 2 performs GRPO-based policy learning, incorporating domain vocabulary construction with proposed five-tier fuzzy matching to handle linguistic variability, along with a three-component reward function (Format, Answer, Reasoning) to guide optimization. Group-relative advantage normalization is further applied to enable stable policy updates.
}
\Description{A two-stage framework diagram. Stage 1 shows automated reasoning data generation using a VLM generator and LLM-based filtering. Stage 2 shows GRPO-based reinforcement learning with domain vocabulary and reward function components.}
\label{fig:framework}
\end{figure*}

\section{Related Work}
\label{sec:related_work}

Agricultural Vision-Language Models.
Recent advances in VLMs have driven the development of domain-specific adaptations for agricultural disease diagnosis~\cite{zhou2024few,awais2025agrogpt,arshad2025leveraging}. Existing studies generally follow two main paradigms. The first focuses on compact model design, as exemplified by~\citet{cao2025small}, who employ image-text contrastive learning for few-shot crop disease identification. The second integrates domain-specific knowledge to enhance representational alignment, as demonstrated by~\citet{yao2024multimodal} through the incorporation of meteorological indicators for multimodal drought detection. In addition, large-scale data initiatives such as AGBase-2000K have facilitated richer knowledge integration via comprehensive multimodal agricultural corpora~\cite{gauba2025agmmu}. Despite these advances,~\citet{liu2024multimodal} reveal that models trained solely via SFT remain prone to substantial performance degradation under domain shifts, highlighting the limited robustness and interpretability of current approaches in open-ended agricultural VQA.

\paragraph{Chain-of-Thought for Interpretability.}

The inherent ``black-box'' nature of VLMs poses a fundamental barrier to their adoption in high-stakes applications such as agriculture, where transparent and trustworthy decision-making is essential~\cite{sun2022black,bommasani2021opportunities,martin2024xai}. To address this limitation, Chain-of-Thought (CoT) prompting~\cite{wei2022chain} has emerged as a prominent technique for improving model interpretability by eliciting explicit, step-by-step reasoning paths. Subsequent research has sought to enhance the reliability of CoT; for instance, self-consistency~\cite{wang2022self} improves robustness by aggregating predictions across multiple reasoning trajectories. However, a critical bottleneck persists: the manual curation of high-quality, domain-specific CoT demonstrations remains prohibitively expensive and difficult to scale~\cite{wang2025beyond,lightman2023letsverifystepstep,kim2023cotcollectionimprovingzeroshot}. This challenge is particularly acute in agriculture, where expert knowledge is required to validate the correctness and relevance of diagnostic reasoning chains, underscoring the pressing need for scalable, automated solutions for CoT generation.

\paragraph{Reinforcement Learning for VLM Post-Training.}

RL offers a powerful paradigm for aligning models with desired behaviors through reward-based feedback, serving as a compelling alternative to supervised fine-tuning by emphasizing outcome-driven optimization~\cite{christiano2017deep,ladosz2022exploration}. Building upon Proximal Policy Optimization (PPO)~\cite{schulman2017proximal}, GRPO~\cite{shao2024deepseekmath} simplifies the training architecture by replacing the learned value function with group-based advantage estimation. This design substantially reduces computational cost while preserving stable convergence. GRPO has demonstrated strong reasoning capabilities in mathematics~\cite{shao2024deepseekmath} and coding~\cite{guo2025deepseek}. Similarly, in specialized domains such as medical vision-language understanding, RL has been successfully adapted to address data scarcity and enhance cross-modal generalization, as demonstrated in recent work such as \citet{zhi2025medgr}.

The application of RL to open-ended agricultural VQA remains an underexplored challenge, with no prior work adapting GRPO to this domain. Unlike medical or mathematical tasks, agricultural diagnosis requires interpretable reasoning under conditions of linguistic diversity, data scarcity, and domain shifts. Existing RL methods typically rely on binary or multiple-choice rewards, which are ill-suited for evaluating free-form, agriculturally grounded explanations. To the best of our knowledge, this work presents the first GRPO-based framework for agricultural VQA. By introducing a domain-aware reward design and automated reasoning data synthesis, the proposed framework jointly improves accuracy, generalization, and interpretability without requiring expert annotations.

\section{Methodology}
\label{sec:methodology}

Figure~\ref{fig:framework} presents the framework. It consists of two stages: a Generative Reasoning Enhancement Engine for constructing a high-quality reasoning dataset, followed by a GRPO Reinforcement Learning stage for training a robust policy with domain-specific rewards.

\subsection{Generative Reasoning Enhancement}
\label{sec:cot_augmentation}

To enable interpretable reasoning without manual annotation, we adopt the three-stage pipeline illustrated in Figure~\ref{fig:framework} (Stage~1).
First, in the \textbf{data processing} stage, input images are resized to a uniform resolution.
Second, for \textbf{reasoning data generation}, a state-of-the-art VLM produces structured reasoning chains in the format
$\langle\text{think}\rangle R \langle/\text{think}\rangle \langle\text{answer}\rangle A \langle/\text{answer}\rangle$.
Third, \textbf{quality filtering} is performed using an LLM-based judge, which evaluates each reasoning chain on a 10-point rubric and retains only those with a score of at least $\tau = 8.0$.
Subthreshold outputs are regenerated via feedback-guided prompting~\cite{stan2025learning}, with up to two refinement attempts.
This stringent filtering process admits only $19\%$ of the initial candidates, thereby ensuring a high density of training signals.

\paragraph{Evaluation Rubric.}
Table~\ref{tab:cot_quality} presents the five-criterion rubric employed to assess the quality of reasoning chains. Each criterion is scored on a scale of 0 to 2, and chains achieving a total score of at least 8 are retained for subsequent GRPO training. This rubric grounds the filtering process in domain-relevant, auditable criteria, prioritizing reasoning quality over stylistic fluency.
A key feature of our design is the dual-model architecture, comprising a generator VLM and an independent LLM-based judge. This separation ensures cross-architectural independence and mitigates the risk of correlated biases that arise when a single model is tasked with both generating and evaluating its own outputs.


\begin{table}[t]
\centering
\small
\begin{tabular}{lcp{4.8cm}}
\hline
\textbf{Criterion} & \textbf{Score} & \textbf{Evaluation Focus} \\
\hline
Accuracy    & 0--2 & Correct plant/ disease ID; no hallucination \\
Completeness & 0--2 & Key elements: plant, symptoms, disease \\
Detail       & 0--2 & Measurements, colors, distribution \\
Relevance    & 0--2 & Diagnosis-relevant; no redundancy \\
Clarity      & 0--2 & Professional terms; logical flow \\
\hline
\textbf{Total} & \textbf{0--10} & \textbf{Threshold: $\tau = 8.0/10.0$} \\
\hline
\end{tabular}
\caption{LLM-based quality evaluation rubric for reasoning.}
\label{tab:cot_quality}
 \vskip -2em
\end{table}

\subsection{GRPO Reinforcement Learning}
\label{sec:grpo}

\subsubsection{Group Relative Policy Optimization}

GRPO optimizes the policy $\pi_\theta$ using group-based advantage estimation~\cite{zheng2025group}, without requiring a separate reward model. For each input $(I, q)$, we sample $G$ responses:

\begin{equation}
o_i \sim \pi_\theta(\cdot \mid I, q), \quad i = 1, \ldots, G
\label{eq:sample}
\end{equation}

\noindent where $o_i$ is a candidate response, $I$ the input image, and $q$ the question. Each response receives a scalar reward $r_i$ from our reward function (Section~\ref{sec:reward}).

\noindent The group relative advantage normalizes rewards within each group of responses to stabilize learning:

\begin{equation}
\begin{aligned}
A_i &= \frac{r_i - \mu_G}{\sigma_G + \epsilon}, \quad
\mu_G = \frac{1}{G}\sum_{j=1}^G r_j, \\
\sigma_G &= \sqrt{\frac{1}{G}\sum_{j=1}^G (r_j - \mu_G)^2}
\end{aligned}
\label{eq:advantage}
\end{equation}

\noindent where $A_i$ is the advantage for candidate $i$, $\mu_G$ and $\sigma_G$ are the group's mean and standard deviation, and $\epsilon$ is a small constant for stability. This normalization helps the model learn from relative quality differences within each group.

The GRPO objective balances policy improvement with KL regularization:

\begin{equation}
\begin{aligned}
\mathcal{J}_G(\theta) &= \mathbb{E}_{(I,q)\sim\mathcal{D}} \Bigg[
\frac{1}{G} \sum_{i=1}^G \min\big( \rho_i A_i, \\
&\quad \text{clip}(\rho_i, 1-\varepsilon, 1+\varepsilon) A_i \big) \\
&\quad - \beta \cdot D_{KL}(\pi_\theta \| \pi_{ref}) \Bigg]
\end{aligned}
\label{eq:grpo_objective}
\end{equation}

\noindent where \(\mathcal{J}_G\) is the GRPO objective; \(\rho_i\) is the probability ratio between current and old policies; the clipping operation and $D_{KL}$ penalty enforce conservative policy updates.

\subsubsection{Reward Function Design}
\label{sec:reward}

A key challenge in agricultural VQA is designing reward functions for open-ended responses with high linguistic diversity~\cite{qian2025toolrl,eschmann2021reward,liu2024multimodal,lai2025med,pan2025medvlm}. We construct domain-specific vocabularies $\mathcal{V}_p$ and $\mathcal{V}_d$ for synonym recognition, then define a three-component reward function:

\begin{equation}
R(o) = w_f R_f(o) + w_a R_a(o) + w_r R_r(o)
\label{eq:total_reward}
\end{equation}

\noindent where $o$ is the candidate response; $R_f$, $R_a$, and $R_r$ denote Format, Answer Exact Match, and Reasoning Quality rewards respectively; $w_f=0.5$ (17\%), $w_a=2.0$ (67\%), and $w_r=0.5$ (17\%) are the component weights; and $R(o) \in [0, 3.0]$ is the total reward.

\paragraph{Domain Vocabularies.}
We construct domain-specific vocabularies $\mathcal{V}_p$ (plant species) and $\mathcal{V}_d$ (disease types) from the benchmark's major crop and disease categories. Each entry includes canonical names, scientific nomenclature (e.g., ``tomato'' $\leftrightarrow$ ``\emph{Solanum lycopersicum}''), and colloquial variations to handle linguistic diversity in agricultural diagnosis.

\paragraph{Format Reward.}
This component ensures structured output with required tags and quality metrics:

\begin{equation}
R_f(o) = \begin{cases}
\sum_{c \in C_f} w_c \cdot r_c(o) & \text{if tags exist} \\
0 & \text{otherwise}
\end{cases}
\label{eq:format_reward}
\end{equation}

\noindent where $C_f = \{\text{struct.}, \text{steps}, \text{content}, \text{length}, \text{quality}\}$ evaluates basic structure with $\langle\text{think}\rangle$ (reasoning) $\langle/\text{think}\rangle$ and $\langle\text{answer}\rangle$ (response) $\langle/\text{answer}\rangle$ tags ($w=0.15$), step structure and content quality ($w=0.15, 0.10$), and appropriate think/answer lengths ($w=0.05$ each), summing to 0.5.

\paragraph{Answer Keyword Reward.}
This component evaluates diagnostic accuracy using the domain vocabularies. For diagnostic questions, we employ weighted dual matching:

\begin{equation}
\begin{aligned}
R_a^{\text{diag}}(o) &= w_p \cdot M_p(o, a) \\
&\quad + w_d \cdot M_d(o, a)
\end{aligned}
\label{eq:answer_diag}
\end{equation}

\noindent where $w_p=0.8$ and $w_d=1.2$ weight plant and disease matching; $M_p(o, a)$ and $M_d(o, a)$ measure matches via five-tier fuzzy scoring ranging from exact synonym match (1.0) to weak relevance (0.25).


For prevention/control questions, we match against method categories:

\begin{equation}
R_a^{\text{ctrl}}(o) = \sum_{c} w_c \cdot \mathds{1}[\text{Contains}(o, \mathcal{V}_c)]
\label{eq:control_reward}
\end{equation}

\noindent where $c \in \{ch, cu, b, t\}$ denotes chemical ($w=0.6$), cultural ($w=0.5$), biological ($w=0.5$), and timing ($w=0.4$) methods; $\mathcal{V}_c$ are category vocabularies; $\mathds{1}[\cdot]$ indicates keyword presence.

\paragraph{Reasoning Quality Reward.}
This component evaluates Chain-of-Thought quality through three dimensions:

\begin{equation}
R_r(o) = \sum_{d \in D_r} w_d \cdot r_d(o)
\label{eq:reasoning_reward}
\end{equation}

\noindent where $D_r = \{\text{logic.}, \text{prof.}, \text{comp.}\}$ evaluates logical coherence th-rough causal patterns (e.g., ``observe...because'') and step connections ($w=0.25$), professional terminology usage in appropriate diagnostic context ($w=0.15$), and reasoning chain completeness covering observation$\rightarrow$analysis$\rightarrow$conclusion flow ($w=0.10$).

\paragraph{Dynamic Evaluation.}

Our reward function dynamically adapts its evaluation criteria according to question type, employing distinct scoring formulations for diagnostic queries (Equation~\ref{eq:answer_diag}) and control/prevention questions (Equation~\ref{eq:control_reward}). To accommodate the inherent lexical variation in open-ended agricultural responses, the function integrates a five-tier fuzzy matching mechanism that spans exact matches to weak semantic relevance~\cite{reichard2025open}. The complete pipeline yields a bounded scalar reward \(r_i \in [0, 3.0]\), which is subsequently used for GRPO optimization (Equation~\ref{eq:grpo_objective}).

\begin{table*}[t]
\centering
\resizebox{0.92\textwidth}{!}{
 \renewcommand\tabcolsep{18pt}
\begin{tabular}{c|c|cc|c}
\hline
\textbf{Model} & \textbf{Method} & \textbf{Crop} & \textbf{Disease} & \textbf{Knowledge} \\
 &  & \textbf{Acc. (\%)} & \textbf{Acc. (\%)} & \textbf{QA (/100)} \\
\hline
\multicolumn{5}{c}{\textbf{Baseline ~\cite{lu2024application}}} \\
\hline
Qwen-VL-Chat (7B) & Zero-shot & 28.40 & 5.00 & 41.0 \\
Qwen-VL-Chat-AG* (7B) & SFT (Frozen encoder) & 84.40 & 66.10 & 88.5 \\
Qwen-VL-Chat-AG (7B)& SFT (Unfrozen encoder) & \textbf{97.40} & \textbf{91.50} & 84.0 \\
\hline
\multicolumn{5}{c}{\textbf{Baseline ~\cite{zhang2025cpjexplainableagriculturalpest}}} \\
\hline
\multirow{3}{*}{Qwen-VL-Chat (7B)} & Expl. Caption & 29.30 & 12.10 & 46.5 \\
 & \quad+Few-shot & 53.39 & 24.49 & 50.0 \\
 & \qquad+Judge & 54.90 & 25.39 & 51.0 \\
\cline{1-5}
\multirow{4}{*}{GPT-5-Nano} & Zero-shot & 47.00 & 11.00 & 65.0 \\
 & Expl. Caption & 60.30 & 31.60 & 84.0 \\
 & \quad+Few-shot & 58.90 & 29.80 & 76.0 \\
 & \qquad+Judge & 63.38 & 33.70 & \textbf{84.5} \\
\hline
\multicolumn{5}{c}{\textbf{Our Methods}} \\
\hline
\multirow{5}{*}{Qwen2.5-VL-3B-Instruct} & Zero-shot & 28.41 & 4.84 & 27.5 \\
 & Few-shot & 36.56 & 6.96 & 45.5 \\
 & SFT & 90.97 & 58.84 & 63.0 \\
 & GRPO & 92.33 & 69.43 & 72.49 \\
 & \textbf{Agri-R1 (Reasoning-Enhanced)} & \textbf{92.58} & \textbf{75.30} & \textbf{84.0} \\
\hline
\textit{GRPO Gain (vs SFT)} & & \textcolor{blue}{+1.36} & \textcolor{blue}{+10.59} & \textcolor{blue}{+9.49} \\
\textit{Reasoning Contribution} & & \textcolor{blue}{+0.25} & \textcolor{blue}{+5.87} & \textcolor{blue}{+11.51} \\
\textit{Total Gain vs SFT} & & \textcolor{blue}{+1.61} & \textcolor{blue}{+16.46} & \textcolor{blue}{+21.0} \\
\textit{Relative Gain} & & \textcolor{blue}{+1.8\%} & \textcolor{blue}{+27.9\%} & \textcolor{blue}{+33.3\%} \\
\hline
\end{tabular}
}
\caption{
Performance comparison on CDDMBench. Baselines include zero-shot and SFT variants of Qwen-VL-Chat (7B)~\cite{lu2024application} as well as prompt-based methods using Qwen-VL-Chat (7B) and GPT-5-Nano~\cite{zhang2025cpjexplainableagriculturalpest}. Our 3B models are trained with GRPO (answer-only rewards) and Agri-R1 (Reasoning-Enhanced with explicit diagnostic reasoning). GRPO yields substantial gains, while explicit reasoning delivers further improvements, particularly on knowledge-intensive tasks.
}
\label{tab:cddmbench_comparison}
 \vskip -1em
\end{table*}

\section{Experiments}
\label{sec:experiments}

\subsection{Datasets and Evaluation}
\label{sec:datasets}
We construct our training datasets from CDDMBench~\cite{liu2024multimodal}, a large-scale agricultural VQA benchmark containing approximately 1.05 million samples across 16 crop species and 60 disease categories. The SFT training set utilizes the full CDDMBench dataset in its standard VQA format. For GRPO training, we apply stratified sampling to obtain 200K samples (19\% of the original corpus), while strictly preserving the original class distribution across crops and diseases. This ratio aligns with reward-guided data efficiency findings~\cite{zhi2025medgr}. For reasoning synthesis, we explore two Generative Reasoning Enhancement Engines (DeepSeek-VL2~\cite{wu2024deepseek} and Qwen2.5-VL-72B~\cite{bai2025qwen2}) paired with four judge configurations (No Judge, self-judge, Qwen2.5-VL-72B, and GPT-4), yielding eight pipeline combinations whose impact on GRPO performance is analyzed in Section~\ref{sec:ablation_pipeline}.

\textbf{Evaluation Protocol: }
Following CDDMBench protocol, we evaluate on: (1) In-distribution test set (3,963 samples) using keyword matching accuracy for crop/disease recognition; (2) Disease Knowledge QA (20 samples) scored by GPT-4 (0-10 scale) on professionalism, completeness, and practicality, following~\cite{liu2024multimodal}; (3) We also evaluate on AgMMU benchmark (770 samples)~\cite{gauba2025agmmu} for cross-scenario generalization using harmonic mean across five subtasks. To complement automatic metrics, Section~\ref{sec:human_eval} further provides expert human evaluation of reasoning quality on a 200-sample subset.

\subsection{Training Configuration}

We adopt Qwen2.5-VL-3B-Instruct~\cite{bai2025qwen2} as our base VLM. Training is conducted on 4 NVIDIA A800 80GB GPUs with DeepSpeed ZeRO-3 optimization. The hyperparameters include a batch size of 160, AdamW optimizer with learning rate $8\times10^{-7}$ and cosine schedule warmup, gradient clipping at 0.3, and BF16 mixed precision. The model is trained for 3 epochs, with the optimal checkpoint selected at step 1,800. For GRPO training, we sample $K=3$ candidate responses per query with temperature $T=0.7$; this group size balances response diversity against computational cost, providing sufficient within-group variance for stable advantage estimation without the memory overhead of larger groups. Throughout training, the KL divergence stabilized between 0.036 and 0.040, confirming that the policy updates remained conservative and did not result in collapse. The entire training process took 98 hours.

\subsection{Baselines}

We evaluate our method against the following baselines: Zero-shot uses the pretrained Qwen2.5-VL-3B-Instruct model with only task prompts. Few-shot augments zero-shot with 5 in-context examples. SFT applies supervised fine-tuning on the complete CDDMBench dataset (1.05M samples). GRPO optimizes with answer correctness rewards only, without explicit reasoning. Agri-R1 (Reasoning-Enhanced) (Ours) is our complete two-stage framework, integrating automated reasoning data synthesis and reasoning-aware reward functions.
Furthermore, we compare our results to published baselines: CDDMBench~\cite{lu2024application} uses Supervised Fine-Tuning on crop disease datasets.
CPJ~\cite{zhang2025cpjexplainableagriculturalpest} is a training-free approach utilizing explainable captions and LLM-as-Judge evaluation.

\subsection{Main Results}
\label{sec:results}
\paragraph{\textbf{Overall Performance on CDDMBench.}}

Table~\ref{tab:cddmbench_comparison} presents comprehensive results comparing our approach with published baselines. Throughout this section, \textit{Agri-R1 (Reasoning-Enhanced)} denotes the complete \textbf{Agri-R1} framework and is used interchangeably with ``Agri-R1'' for brevity. Our Agri-R1 (Reasoning-Enhanced) uses the DeepSeek-VL2 + GPT-4 pipeline (see Section~\ref{sec:ablation_pipeline}).

We highlight three key observations:

(1) \textbf{Crop Recognition}: Agri-R1 achieves 92.58\% accuracy, yielding a +1.61\,pp absolute gain over SFT (90.97\%), with GRPO contributing +1.36\,pp and explicit reasoning adding +0.25\,pp.

(2) \textbf{Disease Recognition}: Agri-R1 reaches 75.30\% accuracy, delivering a +27.9\% relative improvement over SFT. GRPO provides the dominant gain (+10.59\,pp), while explicit reasoning further enhances fine-grained symptom differentiation (+5.87\,pp).

(3) \textbf{Knowledge QA}: Agri-R1 attains 84.0 points, matching state-of-the-art proprietary models with a +33.3\% relative gain over SFT. Notably, explicit reasoning contributes more (+11.51 points) than GRPO alone (+9.49 points), underscoring its critical role in multi-step knowledge integration.

Across all three tasks, the benefit of explicit reasoning increases in the order Crop $<$ Disease $<$ KQA, reflecting each task's growing reliance on causal, multi-step inference rather than pure visual matching.

\begin{figure}[!t]
\centering
\includegraphics[width=0.96\columnwidth]{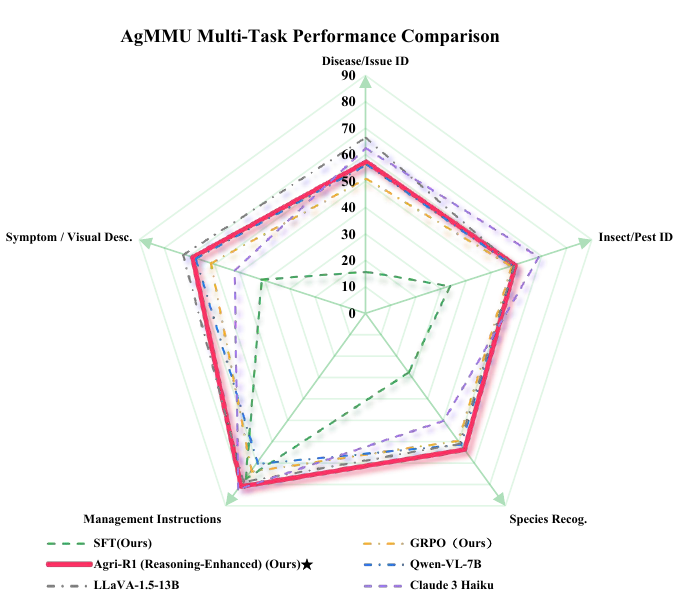}
\caption{AgMMU cross-domain performance. \mbox{Agri-R1 (Reasoning-Enhanced)} (red) outperforms SFT (blue) and GRPO (green) on all five tasks.
}
\Description{A radar chart comparing Agri-R1 (Reasoning-Enhanced) (red) and SFT (blue) across five AgMMU subtasks. The red radar polygon is consistently larger than the blue polygon, particularly on visual tasks.}
\label{fig:agmmu_radar}
 \vskip -1em
\end{figure}

\paragraph{\textbf{Generalization on AgMMU-MCQs.}}
We evaluate cross-domain generalization on AgMMU-MCQs, a challenging subset of the AgMMU benchmark spanning five agricultural reasoning tasks. Using only 3B parameters, Agri-R1 (Reasoning-Enhanced) achieves 66.10\% accuracy, matching LLaVA-1.5-13B (66.73\%) while outperforming Qwen-VL-7B (62.34\%) and Claude 3 Haiku (62.00\%). Figure~\ref{fig:agmmu_radar} visualizes performance across all tasks. SFT experiences a drastic 50.97-point drop from 90.97\% (CDDMBench) to 40.00\% (AgMMU-MCQs). In contrast, GRPO (without explicit reasoning) maintains 59.75\% despite using identical 19\% training data, and Agri-R1 further boosts generalization by +26.10 points. The largest gains appear on Management Instructions and Symptom/Visual Description, where structured reasoning chains supply procedural scaffolding and enable effective integration of visual cues with domain knowledge. These results highlight RL's superior capacity to learn transferable, domain-invariant representations compared with standard supervised fine-tuning.



\section{Analysis}
\subsection{Generator-Judge Configuration Ablation}
\label{sec:ablation_pipeline}

We compare two generators, DeepSeek-VL2 and Qwen2.5-VL-72B, under four judge settings. The results on crop recognition, disease recognition, and knowledge QA are reported in Table~\ref{tab:judge_ablation}. Two key findings emerge:
First, external judge quality is paramount. GPT-4 judging consistently outperforms self-judging and weaker alternatives, revealing a systematic failure mode where generator-judge correlation allows flawed reasoning to survive filtration. This confirms that judge capability, not merely diversity, determines filtering rigor.
Second, generator choice involves task-specific trade-offs. While Qwen2.5-VL-72B excels at fine-grained visual recognition, DeepSeek-VL2 produces more compact, regular reasoning chains that transfer effectively to smaller student models.
Consequently, we adopt DeepSeek-VL2 with GPT-4 judging for main experiments, validating that strict reasoning filtration via high-quality external judges substantially improves downstream inference.

\begin{table}[t]
\centering
\small
\begin{tabular}{llccc}
\hline
\textbf{Generator} & \textbf{Judge} & \textbf{Crop} & \textbf{Disease} & \textbf{KQA} \\
 &  & \textbf{Acc. (\%)} & \textbf{Acc. (\%)} & \textbf{(/100)} \\
\hline
\multirow{4}{*}{DeepSeek-VL2}
 & No Judge & 92.4 & 70.54 & 75.5 \\
 & DeepSeek Judge & 92.45 & 72.5 & 77.0 \\
 & Qwen72B Judge & 92.5 & 73.0 & 81.0 \\
 & GPT-4 Judge & \textbf{92.58} & \textbf{75.3} & \textbf{84.0} \\
\hline
\multirow{4}{*}{Qwen2.5-VL-72B}
 & No Judge & 92.6 & 71.45 & 73.2 \\
 & DeepSeek Judge & 92.9 & 72.6 & 75.0 \\
 & Qwen72B Judge & 93.1 & 72.5 & 74.5 \\
 & GPT-4 Judge & \textbf{94.2} & 74.2 & 81.0 \\
\hline
\end{tabular}
\caption{Generator--Judge ablation across all three metrics. The DeepSeek-VL2 + GPT-4 pipeline generates the highest-quality reasoning text, yielding the strongest overall reasoning performance. This confirms that optimized reasoning synthesis substantially improves downstream inference capability.
}
\label{tab:judge_ablation}
\vskip -1.5em
\end{table}

\begin{table}[t]
\centering
\small
\begin{tabular}{lcccr}
\hline
\textbf{Method} & \textbf{Diag.} & \textbf{Reasoning} & \textbf{Utility} & \textbf{H--G4 $r$} \\
 & \textbf{Acc.} & \textbf{Validity} & & \\
\hline
SFT              & 6.5 & 3.2 & 6.1 & 0.82 \\
GRPO             & 7.6 & 5.6 & 7.3 & 0.86 \\
\textbf{Agri-R1} & \textbf{8.1} & \textbf{7.8} & \textbf{8.0} & \textbf{0.89} \\
\textbf{(Reasoning-Enhanced)} & & & & \\
\hline
\end{tabular}
\caption{Expert evaluation (N=200, scores 0--10). H--G4 $r$ denotes the Pearson correlation between human and GPT-4.
}
\label{tab:expert_eval}
\vskip -1em
\end{table}

\begin{table*}[t]
\centering
\resizebox{.94\textwidth}{!}{
\renewcommand\tabcolsep{19pt}
\begin{tabular}{lrccr}
\toprule
\textbf{Crop} & \textbf{Freq.} & \textbf{SFT} & \textbf{Agri-R1 (Reasoning-Enhanced)} & \textbf{+FA Weight\textsuperscript{$\dagger$}} \\
\midrule
\multicolumn{5}{l}{\textit{High-freq. (>5\%) -- Stable ($\sigma$=3.2\,pp)}} \\
Tomato & 37.19\% & 90.95\% & \textbf{96.05\%} (+5.10) & 95.68\% ($-$0.37) \\
Apple & 29.48\% & 90.94\% & \textbf{97.69\%} (+6.75) & 97.18\% ($-$0.51) \\
Corn & 8.35\% & 91.12\% & \textbf{96.55\%} (+5.43) & 95.86\% ($-$0.69) \\
\midrule
\multicolumn{5}{l}{\textit{Mid-freq. (2-5\%) -- Moderate ($\sigma$=8.7\,pp)}} \\
Potato & 4.21\% & 90.88\% & \textbf{94.23\%} (+3.35) & 93.91\% ($-$0.32) \\
Grape & 3.31\% & 90.84\% & \textbf{100.00\%} (+9.16) & 100.00\% (=) \\
Soybean & 3.15\% & 91.05\% & \textbf{93.87\%} (+2.82) & \textbf{95.08\%} (+1.21) \\
\midrule
\multicolumn{5}{l}{\textit{Low-freq. (<2\%) -- High Variance ($\sigma$=22.1\,pp)}} \\
Bell Pepper & 1.73\% & 91.14\% & 83.54\% (\textcolor{red}{-7.60}) & \textbf{89.75\%} (\textcolor{blue}{+6.21}) \\
Raspberry   & 1.61\% & 100.00\% & 80.00\% (\textcolor{red}{-20.00}) & \textbf{89.47\%} (\textcolor{blue}{+9.47}) \\
Cherry      & 1.23\% & 91.30\% & 31.88\% (\textcolor{red}{-59.42}) & \textbf{52.10\%} (\textcolor{blue}{+20.22}) \\
\bottomrule
\end{tabular}}
\caption{Crop recognition by training frequency. Standard GRPO remains stable on high-frequency crops ($\sigma$=3.2\,pp) but becomes highly variable on low-frequency crops ($\sigma$=22.1\,pp), where all classes below 2\% regress. FA weighting restores all low-frequency crops to at or above their SFT baselines. \textsuperscript{$\dagger$}For mid-frequency crops ($f(c)\geq2.5\%$), FA weighting causes only minor changes ($\leq$1.2\,pp), indicating that its main effect is concentrated on the long tail.}
\label{tab:crop_analysis}
\vskip -0.5em
\end{table*}

\subsection{Expert Evaluation of Reasoning Quality}
\label{sec:human_eval}

Protocol.
To complement automatic metrics, which focus on correctness but do not fully reflect real-world reliability or practical utility, we conducted a human evaluation on 200 randomly sampled responses from the CDDMBench test set (150 diagnostic questions and 50 knowledge QA questions). Two agricultural experts, blind to model identity, independently scored the outputs using the same 0--10 rubric as in Table~\ref{tab:cot_quality} across three dimensions: Diagnostic Accuracy, Reasoning Validity, and Practical Utility. Inter-annotator agreement was strong ($\kappa=0.84$), and averaged scores were used in the final analysis.

\paragraph{Results.}
Table~\ref{tab:expert_eval} summarizes the human evaluation results. Agri-R1 (Reasoning-Enhanced) achieves the highest scores across all three dimensions and the strongest alignment with GPT-4 ratings ($r=0.89$). The largest improvement appears in Reasoning Validity: SFT scores only 3.2/10 (lacking explicit structure), standard GRPO reaches 5.6/10 (still largely implicit), while Agri-R1 attains 7.8/10 thanks to explicit \texttt{<think>} supervision. This substantial gain aligns directly with the core design of our framework. Notably, human ratings correlate strongly with GPT-4 scores across all methods ($r \ge 0.82$), confirming that our LLM-based scoring reliably approximates expert judgment and supports its use in both data filtering and reward design.



\subsection{Frequency-Induced Bias in Crop Recognition}
\label{sec:bias_analysis}

The crop-level breakdown in Table~\ref{tab:crop_analysis} shows a strong dependence on training frequency. High-frequency crops ($>$5\%) improve in a fairly uniform way, with a standard deviation of 3.2\,pp. The low-frequency group ($<$2\%), in contrast, is much less stable, with a standard deviation of 22.1\,pp. All three classes in this range decline under standard GRPO, and the drop for Cherry is particularly severe, from 91.30\% to 31.88\%.

This pattern is consistent with what one would expect from gradient competition. Frequent classes appear more often, contribute more reward signal, and therefore exert more influence on the update direction. Apple, for example, accounts for 29.48\% of the training data, whereas Cherry accounts for only 1.23\%. Over the course of training, such an imbalance can gradually bias the model toward frequent categories at the expense of rare ones. The same issue reappears in disease recognition and is discussed again in Section~\ref{sec:disease_analysis}. In both cases, the underlying problem is that the group-relative advantage in Equation~\ref{eq:advantage} is not frequency-aware, so batch statistics are dominated by classes that are already well represented.

To test whether this is the main source of the problem, we introduce a frequency-aware (FA) variant that rescales Equation~\ref{eq:total_reward} by $w_{\mathrm{freq}}(c)=1+\alpha\cdot\max(0,\,\theta-f(c))$, where $\alpha=50$ and $\theta=2.5\%$.
 Under this setting, classes with $f(c)\geq2.5\%$ keep weight 1.0, while the rarest classes receive up to 3.35$\times$ reward amplification. The effect is concentrated almost entirely on the long tail. Cherry rises from 31.88\% to 52.10\%, and all three low-frequency crops return to or exceed their SFT baselines. At the dataset level, overall crop accuracy increases from 92.28\% to 93.54\%, which is a 1.37\% relative gain over Agri-R1.

\begin{figure}[!t]
\centering
\includegraphics[width=\columnwidth]{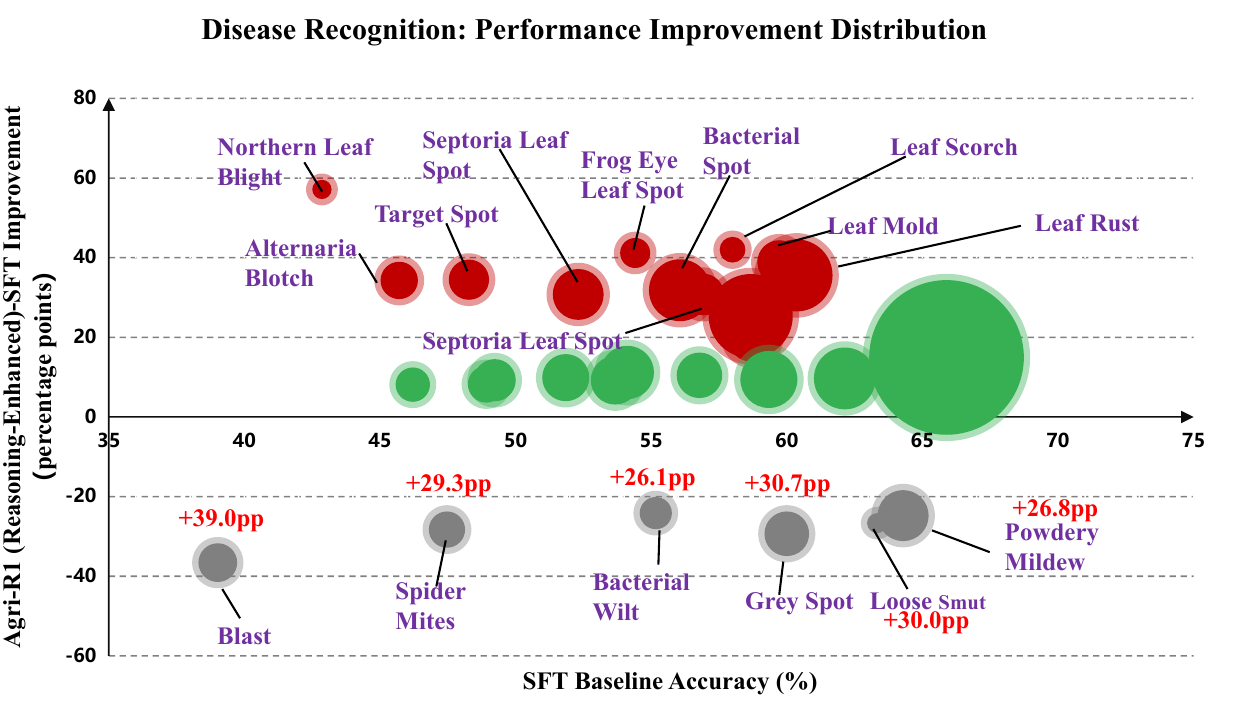}
\caption{Disease recognition improvement by category (\mbox{Agri-R1} vs.\ SFT).
Green points ($>$15\% gain) cluster in the 3--8\% frequency range, while gray points denote low-frequency diseases ($<$2\%) that regress under standard training. Red annotations mark the recovery obtained with FA weighting (\ref{sec:bias_analysis}).}
\Description{A scatter plot showing disease recognition improvements for each disease category. Points are colored green for improvements greater than 15\% and red for declines greater than 15\%. Point size indicates training data proportion.}
\label{fig:disease_improvement}
\vskip -1.5em
\end{figure}

\subsection{Analysis of Fine-Grained Disease Recognition}
\label{sec:disease_analysis}

Figure~\ref{fig:disease_improvement} shows disease-level gains relative to SFT. Improvements concentrate in the 3--8\% frequency range, where the reward signal is stable enough for GRPO to yield consistent gains while SFT leaves room for improvement. Outside this band, performance declines: below 3\%, updates are noisy and gains inconsistent; above 8\%, SFT already performs well. A key concern is the long tail: six diseases with frequencies below 2\% each lose over 20 percentage points under standard training. The KL term in Equation~\ref{eq:grpo_objective} mitigates drift from SFT but does not resolve the imbalance when common and rare classes contribute uneven signal within each batch.

The FA weighting introduced in Section~\ref{sec:bias_analysis} largely addresses this issue. As shown by the annotations in Figure~\ref{fig:disease_improvement}, all six regressed diseases recover to 1--3\,pp above their SFT baselines. This lifts overall disease accuracy from 75.3\% to 78.3\%, which corresponds to a 4.0\% relative improvement over Agri-R1. Most of that gain comes from categories in the sub-2\% tail.

\begin{figure}[!t]
\centering
\includegraphics[width=\columnwidth]{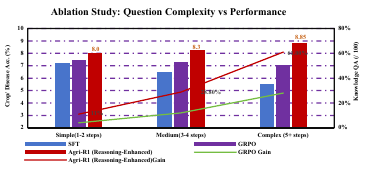}
\caption{Ablation on Disease Knowledge QA (20 samples). Standard GRPO without explicit reasoning (purple) yields consistent but limited gains (+4\% to +28\%), whereas Agri-R1 (Reasoning-Enhanced) (red) achieves larger improvements on difficult questions, reaching +61\%.}
\Description{A bar chart comparing GRPO and Agri-R1 (Reasoning-Enhanced) performance on Disease Knowledge QA questions grouped by complexity level. The reasoning-enhanced approach shows progressively larger gains for more complex questions.}
\label{fig:complexity}
\vskip -1em
\end{figure}

\subsection{Analysis of Reasoning Capability}

\textbf{Scaling of reasoning benefits with task complexity.}The benefit of explicit reasoning varies with task complexity. On simple visual recognition tasks, gains remain modest, as standard GRPO already captures most of the available improvement. The advantage becomes substantially larger on knowledge-intensive questions that require multi-step integration of agricultural knowledge. Figure~\ref{fig:complexity} illustrates this trend: on the most difficult Disease Knowledge QA subset, standard GRPO yields up to +28\% gain, whereas Agri-R1 (Reasoning-Enhanced) reaches +61\%---a 2.2$\times$ larger improvement. These results show that GRPO effectively refines answer distributions, yet explicit reasoning chains are essential for organizing intermediate steps when multi-step inference is required.


\paragraph{\textbf{Multi-Model KQA Progression.}}
The KQA trajectory in Table~\ref{tab:cddmbench_comparison} follows the same trend. Zero-shot performance starts at 27.5. SFT raises it to 63.0, and GRPO without explicit reasoning further improves it to 72.49. The largest single increase then comes from GRPO to Agri-R1 (Reasoning-Enhanced), which adds another 11.51 points. This gain is larger than the improvement from zero-shot to few-shot prompting, and also larger than the gap between SFT and standard GRPO. The final score reaches 84.0, roughly three times the zero-shot baseline. Taken together, these results suggest that explicit reasoning is most valuable on the part of the benchmark where visual recognition is no longer the main difficulty and multi-step knowledge integration becomes the bottleneck.

\begin{figure}[!t]
\centering
\includegraphics[width=1.0\columnwidth]{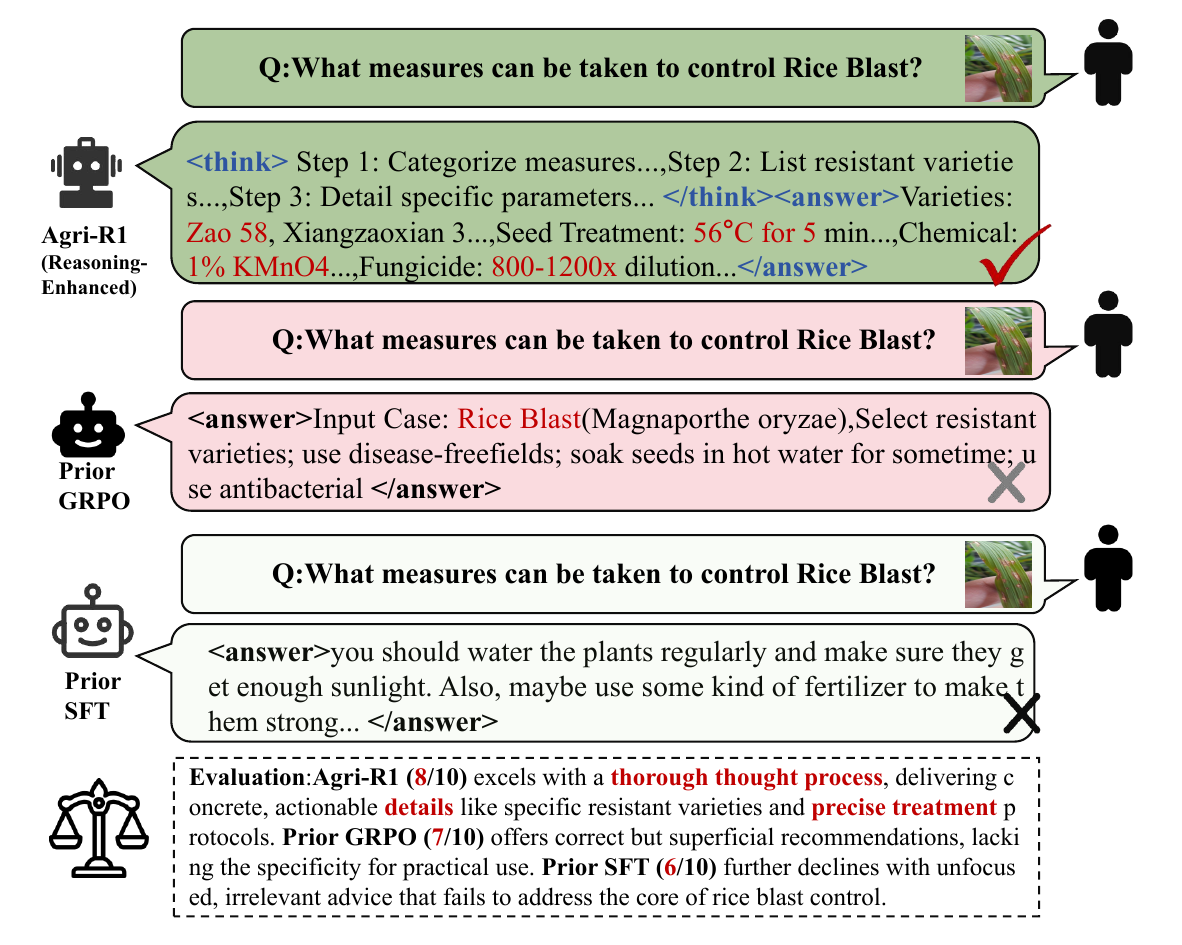}
\caption{Comparison of diagnostic reasoning quality on Rice Blast control. SFT (6/10) provides broad but unfocused advice. GRPO (7/10) identifies correct strategies but lacks operational specificity. Agri-R1 (8/10) produces a structured response with actionable details, including crop varieties, treatment protocols, and dilution ratios.}
\Description{A three-stage comparison showing the progression from Prior SFT (unfocused advice) to Prior GRPO (correct but superficial) to Agri-R1 (structured reasoning with specific parameters). Each stage shows the model's response to a Rice Blast control question with corresponding quality scores.}
\label{fig:cot_example}
\vskip -1.5em
\end{figure}

\paragraph{\textbf{Case Study: Qualitative Analysis of Explicit Reasoning Output.}}
Figure~\ref{fig:cot_example} gives a concrete example with a Rice Blast control question. The SFT model (6/10) produces general advice, but the response is broad and not very specific to the disease. GRPO (7/10) improves the answer and identifies the right intervention direction, yet the output still lacks details that would make it directly actionable. Agri-R1 (8/10) gives a more structured response and includes concrete information such as variety names (Zao 58, Xiangzaoxian 3), treatment conditions (56$^\circ$C for 5 min, 1\% KMnO$_4$), and fungicide dilution ratios (800--1200$\times$). The example reflects the broader pattern in the quantitative results: RL improves correctness, while explicit reasoning more noticeably improves completeness and usefulness.

\section{Conclusion and Future Work}

In this paper, we present \textbf{Agri-R1}, the first GRPO-based framework for open-ended agricultural VQA. By automating reasoning synthesis and designing a domain-aware fuzzy-matching reward, we achieve data-efficient, interpretable, and generalizable disease diagnosis with a compact 3B model that surpasses larger 7B–13B SFT baselines. Our analysis reveals that frequency-induced gradient competition, not model capacity, is the root cause of rare-category collapse under GRPO, challenging the assumption that scaling model size or rewards alone can overcome imbalance. Instead, targeted interventions like frequency-aware reward weighting effectively recover sub-2\% categories, demonstrating that structural imbalances require structural remedies. Beyond agriculture, this insight offers a template for RL-based VLM adaptation in expert-scarce domains such as medical diagnosis and industrial inspection, highlighting that RL effectiveness depends on both algorithmic design and the quality of its guiding data and rewards.

Looking forward, we envision three promising directions: (1) principled solutions to frequency-induced collapse, such as dynamic curriculum learning rewards; (2) temporal modeling via recurrent or state-space architectures to capture disease progression; and (3) richer, multi-turn diagnostic dialogues for more natural farmer-AI interaction. 
\textbf{Agri-R1 establishes that automated reasoning synthesis, paired with domain-aware reward design, can unlock the potential of reinforcement learning for specialized domains, paving the way toward more accessible, interpretable, and generalizable AI in the real world}.

%

\begin{acks}
This work was supported in part by the International Communication Research Project (China International Communications Group) under Grant 25ATILX01; in part by the Young Scientists Fund of the National Natural Science Foundation of China (NSFC) under Grant 62506084; in part by the YoungScientists Fund of the National Natural Science Foundation of China (NSFC) under Grant 32500997; and in part by the Government Special Support Funds for the Guangdong Institute of Intelligence Science and Technology.
\end{acks}

\bibliographystyle{ACM-Reference-Format}
\bibliography{custom_agri_R1}

\end{document}